\documentclass[runningheads]{llncs}
\usepackage[T1]{fontenc}

\usepackage{graphicx} 

\usepackage{amsmath}
\usepackage{amsfonts}
\usepackage{wrapfig}
\usepackage{lipsum}
\usepackage{subcaption}

\usepackage[square,sort,comma,numbers]{natbib}
\title{Prompt-Tuning Bandits: Enabling Few-Shot Generalization for Efficient Multi-Task Offline RL}
\titlerunning{Enabling Few-Shot Generalization for Efficient Multi-Task Offline RL}
\date{}

\author{Finn Rietz$^{\dagger,}$\inst{1,2}\orcidID{0000-0001-8151-4692} \and
Oleg Smirnov$^{\ast,}$\inst{2}\orcidID{0000-0003-1433-7037} \and \\
Sara Karimi$^{\ast,}$\inst{2}\orcidID{0000-0002-0638-7352} \and
Lele Cao$^{\ast,}$\inst{2}\orcidID{0000-0002-5680-9031}}
\authorrunning{F. Rietz et al.}
\institute{\"Orebro University\\
\email{finn.rietz@oru.se}\\
\and
King, Microsoft Gaming\\
\email{\{oleg.smirnov,sarakarimi,lelecao\}@microsoft.com}\\
$^\dagger$ Corresponding author\\
$^\ast$ Equal contribution
}

\begin{document}

\maketitle

\begin{abstract}
    Prompting has emerged as the dominant paradigm for adapting large, pre-trained transformer-based models to downstream tasks. The Prompting Decision Transformer (PDT) enables large-scale, multi-task offline Reinforcement Learning (RL) pre-training by leveraging sto\-chas\-tic trajectory prompts to identify the target task. 
    However, these prompts are sampled uniformly from expert demonstrations, overlooking a critical limitation: \textit{not all prompts are equally informative for differentiating between tasks}. 
    This limits generalization and adaptation, especially in low-data or open-world settings where sample efficiency is crucial.
    To address this issue, we propose a lightweight, inference-time, bandit-based prompt-tuning framework.  The bandit explores and optimizes trajectory prompt selection to enhance task performance, while avoiding costly fine-tuning of the transformer backbone.
    Our experiments indicate not only clear performance gains due to bandit-based prompt-tuning, but also better sample complexity, scalability, and prompt space exploration compared to prompt-tuning baselines.
    These results highlights the importance of adaptive prompt selection mechanisms for efficient generalization in offline multi-task RL.
\end{abstract}

\section{Introduction}
Recent advances in Artificial Intelligence (AI) research have demonstrated the strength of large, pre-trained transformer-based foundation models in many domains, including language \cite{radford2018improving, brown2020language}, vision \cite{radford2021learning, dosovitskiy2020image}, and reinforcement learning \cite{reed2022generalist, li2023survey}. These large models leverage vast and diverse offline datasets to acquire generalizable representations that can solve many downstream tasks.
A prominent strategy for leveraging these models in zero- and few-shot settings involves conditioning them on a \textit{prompt} -- a structured input that specifies the current objective.
By keeping the prompt in context, the model ensures that subsequently generated tokens are aligned with the task. Consequently, the performance of a pre-trained model in a downstream task is contingent not only on the coverage of the pre-training data but also on the quality and informativeness of the provided prompt~\cite{hu2023prompt, lin2023use, lester2021power}.

Building on the success of transformer-based multi-task language models, Offline Reinforcement Learning (ORL) has increasingly adopted transformer architectures, such as the Decision Transformer~(DT)~\cite{chen2021decision}, to address sequential decision-making problems. In the multi-task setting, DT has been extended to the \textit{Prompting Decision Transformer}~(PDT)~\cite{xu2022prompting}, which leverages \textit{stochastic trajectory prompts}, multiple segments of expert demonstrations, to enable task-conditioned pre-training and to facilitate few-shot adaptation. These prompts serve as task descriptors that allow PDT to distinguish tasks and to generate actions aligned with the optimal policy distribution for each task. 
However, PDT samples these prompts uniformly at random from the demonstration dataset, overlooking a crucial limitation: not all prompts are equally informative for identifying the target task, which can lead to performance degradation and hinder generalization capabilities when uninformed sampling methods are employed.
This limitation is especially problematic in applications where online exploration is costly and where adaptation to the target task must happen efficiently and robustly.

Although prompt-tuning is becoming increasingly popular to increase the few-shot performance of Large Language Models (LLMs), e.g. \cite{lester2021power, lin2023use, shi2024best, pmlr-v235-chen24e}, prompt-tuning of RL decision transformers is still underexplored. Hu et al.~\cite{hu2023prompt, hu2024prompt} propose prompt-tuning for PDT by estimating the prompt's gradient with respect to online task return and by incrementally updating the prompt. However, performing gradient ascent directly on the prompt tokens ignores causal structures in the prompt and is not applicable in discrete settings. Yuan et al.~\cite{yuan2024pre} also propose a prompt-tuning approach for PDT, however, their method relies online exploration and additional data collection in the downstream tasks.

Thus, to address the limitations of PDT and prior prompt-tuning works, we introduce a simple yet effective bandit-based prompt-tuning framework that actively explores the prompt space. By formulating prompt selection as a contextual bandit problem, our method systematically identifies and exploits prompts that maximize downstream task performance, without requiring costly modifications to the pre-trained Transformer backbone. This approach is scalable, computationally efficient, and seamlessly integrates with PDT, enhancing performance while eliminating the need for additional task-specific fine-tuning. We summarize our contributions as follows:
\begin{itemize}
    \item We introduce a bandit-based prompt-tuning approach to explore the prompt space at inference time, aiming to find the most effective prompt sequence for each downstream task.
    \item We validate the effectiveness of our approach in a controlled proof-of-concept environment, demonstrating that adaptive prompt selection significantly improves the performance of a pre-trained PDT model.
    \item We benchmark our method against two optimization-based prompt-tuning baselines and evaluate it using two bandit strategies across varying numbers of prompt segments.
    \item We provide analysis on exploration effectiveness and sensitivity to the initialization among different approaches. 
\end{itemize}

Furthermore, ongoing work explores extending this method to more complex environments. Our findings underscore the critical role of prompt optimization in offline RL and reinforce the broader significance of prompt quality in transformer-based decision-making models.

\section{Preliminaries}
This section covers the background of our method. We define our learning objective in Sec.~\ref{sec:orl}, formalize the contextual bandit problem in Sec.~\ref{sec:cmab},
and review the PDT~\cite{xu2022prompting} architecture in Sec.~\ref{sec:pdt}.

\subsection{Offline multi-task RL}\label{sec:orl}
An offline multi-task RL problem consists of a set of training tasks $\mathcal{T}^\text{train}$ and optionally several holdout test tasks $\mathcal{T}^\text{test}$. 
Each task $\mathcal{T}_i \in \{ \mathcal{T}_1, \mathcal{T}_2, \dots, \mathcal{T}_n,\}$ corresponds to a Markov Decision Process (MDP), defined as the tuple $\mathcal{M}_i = \langle \mathcal{S}_i, \mathcal{A}_i, r_i, d_i, \gamma_i, \mu_i^0 \rangle$. 
Here, $\mathcal{S}_i$ is the state space, $\mathcal{A}_i$ is the action space, $r_i: \mathcal{S}_i \times \mathcal{A}_i \to \mathbb{R}$ represents the reward function, $d_i: \mathcal{S}_i \times \mathcal{A}_i \times \mathcal{S}_i \to [0, 1]$ defines the discrete-time transition dynamics, $\gamma_i \in (0, 1]$ is the discount factor, and $\mu_i^0$ is the initial state distribution of MDP $i$. 

For each task $\mathcal{T}_i$, we assume access to an offline trajectory dataset $\mathcal{D}_i$.
The trajectories in $\mathcal{D}_i$ can be collected using one or more policies of arbitrary quality. In addition, for PDT, we require a small set of expert demonstrations $\mathcal{P}_i$ to sample stochastic trajectory prompts from.
Our goal is to exploit the available offline data to compute a generalized policy, $\pi(\mathbf{s}, \rho) \to \mathbf{a}$, capable of solving all tasks in $\mathcal{T}^\text{train}$.
Here, $\rho$ is a task descriptor like an index, one-hot encoding, or prompt, ensuring the policy is aware of the current task.
The learning objective for the generalized policy is to maximize the expected discounted reward objective in Eq.~\eqref{eq:generalized-objective} for each task $\mathcal{T}_i \in \mathcal{T}^\text{train}$.
\begin{equation}\label{eq:generalized-objective}
    J(\pi, i) = \mathbb{E} _{\mathbf{a}_t \sim \pi(\rho), \mathbf{s_t}\sim d_i} \left[ \sum_{t=0}^\infty \gamma_{i}^t r_i(\mathbf{s}_t, \mathbf{a}_t) \right], \rho \sim \mathcal{P}_i
\end{equation}

\subsection{Contextual Multi-Armed Bandits}\label{sec:cmab}
Multi-Armed Bandits (MABs) provide a framework for optimizing stochastic reward functions over the course of $K$ rounds.
For each round $k \in \{ 1, \dots, K\}$ the bandit selects an action $a_k \in \mathcal{A}_b$ by pulling one of its arms, where $\mathcal{A}_b$ denotes the bandit's set of arms.
It then perceives a stochastic reward $r_k \sim R(a_k)$ for performing that action, where $R$ is the reward distribution. 
The goal is to maximize the cumulative reward $\sum_{k=1}^K r_k$ over the $K$ rounds, which requires balancing exploration and exploitation of the available arms while minimizing cumulative regret~\cite{auer2002finite}:
\begin{equation}
    \text{Regret}(K) = \sum_{k=1}^K \left[ \underset{a \in \mathcal{A}_b}{\max}\ \mathbb{E}[R(a)] - \mathbb{E}[r_k] \right]
\end{equation}

Contextual Multi-Armed Bandits (CMABs) extend standard MABs by 
incorporating additional information (i.e. ``context'') $\mathbf{c}_k \in \mathcal{C}$ observed at each round $k$. 
The stochastic reward depends on both the action and the context, $r_k \sim R(a_k \mid \mathbf{c}_k)$, meaning a CMAB's objective is to learn a policy $\pi: \mathcal{C} \to \mathcal{A}_b$ that maximizes the expected reward objective $\mathbb{E}[\sum_{k=1}^K R(\pi(\mathbf{c}_k) \mid \mathbf{c}_k)]$. 
By exploiting the cross-arm features given by the context, CMABs are credited with better sample efficiency and generalization than their non-contextual counterparts~\cite{li2010contextual}, making them well-suited for efficient prompt-tuning.

\subsection{Prompting Decision Transformer}\label{sec:pdt}
Prompting Decision Transformer (PDT)~\cite{xu2022prompting} treats offline multi-task RL as a sequence learning problem by autoregressively modeling the trajectories in the available offline datasets. 
Trajectories consist of $(\hat{r}_t, \mathbf{s}_t, \mathbf{a}_t)$ triplets, with $\hat{r}_t = \sum^T_{t'=t} r_{t'}$ being return-to-go, needed for conditioning on optimal return. 
For all training tasks $\mathcal{T}^\text{train}$, PDT learns to model the sequence $\mathbf{x}$ in Eq.~\eqref{eq:pdt-seq} by autoregressively predicting the action tokens, where $\odot$ denotes concatenation.
The prompt $\rho$ in Eq.~\eqref{eq:pdt} consists of $J$ segments, each of length $H$, that are sampled uniformly from the expert demonstrations $\mathcal{P}_i$ for that task.
\begin{figure*}[]
\begin{equation}
        \rho = \big(
        \overbrace{
        \hat{r}_l, \mathbf{s}_l, \mathbf{a}_l, 
        \dots,
        \hat{r}_{l+H}, \mathbf{s}_{l+H}, \mathbf{a}_{l+H}}
        ^{\text{ $\Tilde{\tau}_1$: segment 1}},
        \dots,
        \overbrace{
        \hat{r}_k, \mathbf{s}_k, \mathbf{a}_k, 
        \dots,
        \hat{r}_{k+H}, \mathbf{s}_{k+H}, \mathbf{a}_{k+H}}
        ^{\text{$\Tilde{\tau}_J$: segment } J}
        \big)
    \label{eq:pdt}
\end{equation}
\begin{equation}
    \mathbf{x} = 
        \big( \rho \big) \odot
        \big(
        \overbrace{
        \hat{r}_{t-N}, \mathbf{s}_{t-N}, \mathbf{a}_{t-N}, 
        \hat{r}_{t-N+1}, \mathbf{s}_{t-N+1}, \mathbf{a}_{t-N+1}
        \dots,
        \hat{r}_{t}, \mathbf{s}_{t}, \mathbf{a}_{t}
        }^{\tau_{N:t}\: N\  \text{most recent transitions}}
        \big)
    \label{eq:pdt-seq}
\end{equation}
\end{figure*}
Instead of relying on uninformed random sampling, we hypothesize that prompts can vary in their usefulness for describing the downstream task, based on segment composition and overlap between demonstrations for multiple tasks, and propose to optimize prompt and segment selection with a CMAB approach. As we detail in the next section, the bandit explores directly in the prompt space and learns to select the best prompt constructible from $\mathcal{P}_i$.
        
\section{Method: Prompt-tuning contextual bandit}\label{sec:method}
We propose a contextual multi-armed bandit (CMAB) architecture to optimize the prompt selection and segment composition to improve the performance of a pre-trained PDT backbone on a downstream task $\mathcal{T}_i$. 
To this end, we assume access to a PDT $\theta^*$, pre-trained until convergence on a multi-task dataset $\mathcal{D}$, a small number of expert demonstrations $\mathcal{P}_i$ to select prompts from, and a simulator $\mathcal{M}_i$ to evaluate online performance for the downstream task $i$.

At a high level, our approach operates as follows. For each round $k \in \{1, \dots, K \}$, the bandit selects a prompt $\rho_k$ from $\mathcal{P}_i$ which is prepended to the PDT's input according to Eq.~\eqref{eq:pdt-seq}.
We then proceed by rolling out the PDT, conditioned on $\rho_k$, in $\mathcal{M}_i$ and take note of the achieved online return $G_k = \sum_{t=0}^T r_i(\mathbf{s}_t, \mathbf{a}_t) \mid \mathbf{a}_t \sim \pi(\mathbf{x}_t, \theta^*)$ for that round. 
Note that while $\tau_{N:t}$ in Eq.~\eqref{eq:pdt-seq} is dynamically updated to reflect the last $N$ steps in the episode, the prompt remains fixed during an entire episode.
From the bandit's perspective, $G_k$ serves as a reward for selecting prompt $\rho_k$, and the tuple $\langle \rho_k, G_k \rangle$ is stored for training the bandit's reward model.

We now detail our CMAB architecture and how it constructs prompts at each round $k$.
A simple but na\"ive approach to bandit based prompt-tuning could be to maintain one arm per prompt constructible from the demonstrations in $\mathcal{P}_i$, however, the search space would grow \textit{combinatorially} with the number of segments $J$ in the prompt under this architecture. 
We instead propose a contextual bandit with only $J$ arms, one for each segment in the prompt, taking the structure of the prompt space and similarity between segments into consideration.
Our bandit maintains a separate reward model $\phi_j: \Tilde{\tau} \to \mathbb{R}$ for each arm $j \in \{ 1, \dots, J\}$, and treats prompt segments as context.
Each of these reward models estimates the return achieved by the PDT for task $i$, when segment $\Tilde{\tau} \in \mathcal{P}_i$ is placed at position $j$ in the prompt. 
Thus, at each round $k$, our bandit predicts the reward for each segment in each position of the prompt, resulting in a prediction matrix $\mathbf{Y}$, with $J$ columns and rows equal to $|\mathcal{P}_i|$, the number of segments in the given expert demonstration dataset.
To select a prompt, the bandit can either exploit based on accumulated knowledge and $\arg \max \mathbf{Y}$ along the segments' dimension, or explore using some exploration mechanisms such as $\epsilon$-greedy, Upper Confidence Bounds (UCB)~\cite{auer2002finite}, or Thompson Sampling~\cite{thompson1933likelihood}. 
To summarize, we propose a bandit architecture tailored to efficient prompt-tuning of PDT models and maintain linear rather then combinatorial scaling with the prompt size.

\section{Experiments}
This section outlines our experimental procedure. We first introduce the multi-task environment and offline dataset, then describe the baselines in Sec.~\ref{sec:baselines} followed by results and analysis in Sec.~\ref{sec:results}.

\textbf{Environment}: We evaluate our proposed prompt-tuning bandit architecture in a 2D proof-of-concept environment. This environment features a planar 2D point agent that has to reach a goal coordinate. The state contains the agent's 2D coordinate at each step $t$. 
The action space contains two continuous actions for translating on the plane, with the step size being limited by projecting the translation vector on a unit circle with a radius of 0.1. In addition, the action space contains a binary \texttt{stop} action which allows the agent to terminate the episode. 
\begin{wrapfigure}{r}{0.35\textwidth}
  \centering
    \includegraphics[width=0.3\textwidth]{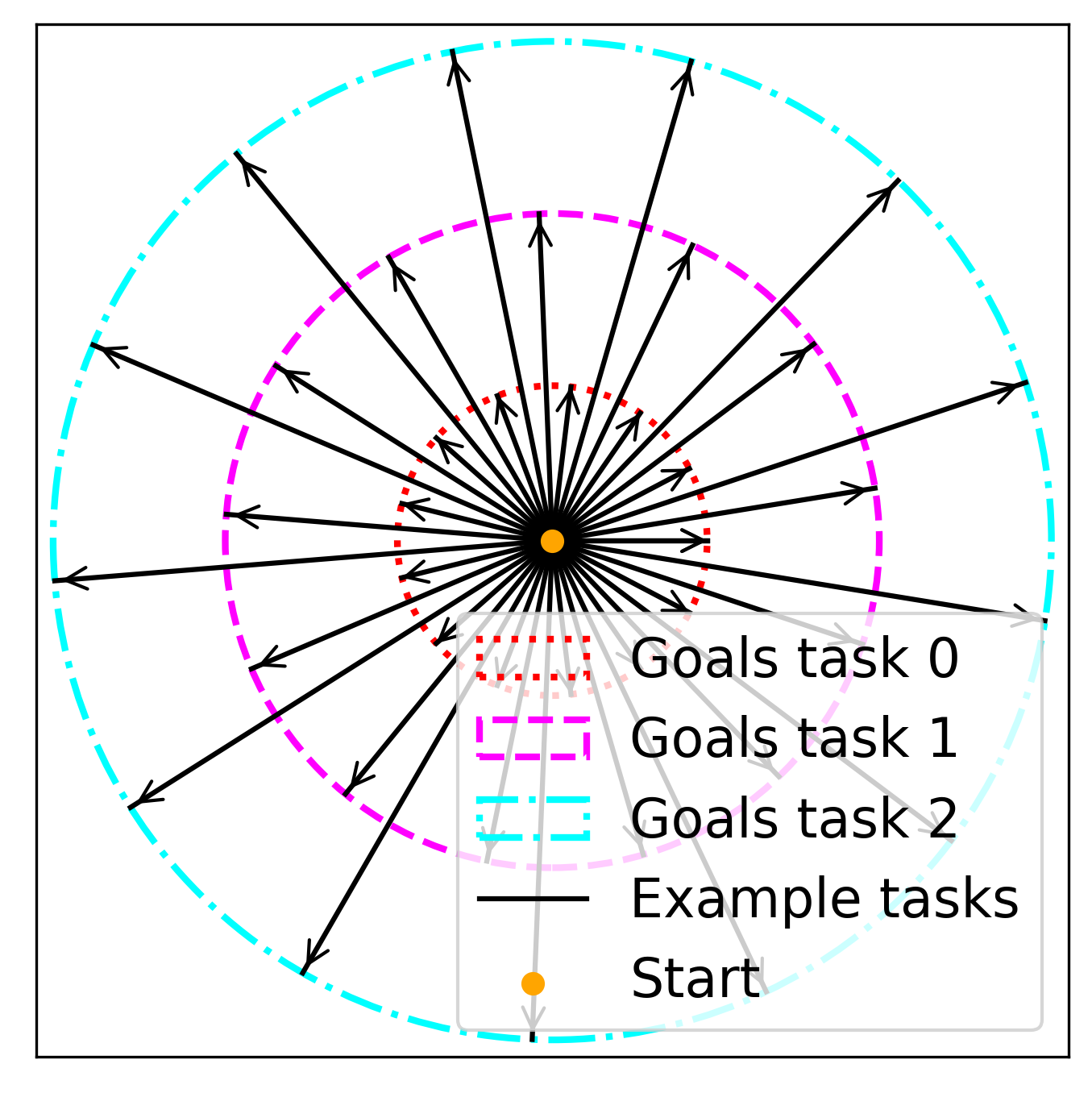}
  \caption{Our proof-of-concept, 2D multi-task environment.}
  \vspace{-15pt}
\end{wrapfigure}
When selected, the episode ends, and the agent receives a sparse reward proportional to its distance from the goal. A bonus of +10 reward is provided (discounted for exceeding the optimal number of steps) for stopping in close proximity of the goal coordinate.
To create a multi-task setting, we parameterize tasks by $(r, \alpha)$, the goal's radius and angle. We discretize the task sparsely using 20 discrete angles $\alpha \in \{ 0.1 \cdot \pi, 0.2 \cdot \pi, \dots, 2 \cdot \pi  \}$ and three discrete radii $r \in \{ 0.9, 1.9, 2.9 \}$, yielding a total of 60 tasks.

\textbf{Offline dataset and pre-training}:
We collect an offline multi-task dataset by training Proximal Policy Optimization (PPO)~\cite{schulman2017proximal} for 1M steps on each of the 60 tasks, storing the trajectories as $\mathcal{D}_i$. We extract trajectories from the top percentile from $\mathcal{D}_i$ to serve as expert demonstrations $\mathcal{P}_i$ for that task.
We then train PDT, without modifications, on $\mathcal{D} = \{ \mathcal{D}_1, \mathcal{D}_2, \dots, \mathcal{D}_{60}\}$ and $\mathcal{P} = \{ \mathcal{P}_1, \mathcal{P}_2, \dots, \mathcal{P}_{60}\}$ until convergence; see \cite{xu2022prompting} for details.

\subsection{Baselines}\label{sec:baselines}
We compare our proposed bandit-based prompt-tuning method, qualitative and empirically, against the following baselines.

\textbf{Standard PDT}~\cite{xu2022prompting} without prompt-tuning: This baselines reveals the possible performance gains due to prompt-tuning at inference time.

\textbf{ZO-RankSGD}-based prompt-tuning~\cite{hu2023prompt}: Closely related to our bandit-based method, this approach proposes prompt-tuning for PDT by employing ZO-RankSGD~\cite{tang2023zeroth} to estimate the gradient of the prompt with respect to online task return $G$. The method samples and initial prompt $\rho_0 \sim \mathcal{P}_i$, and, at each rounds $k$, estimates the gradient 
$\hat{\nabla}_{\rho} G$ 
based on the ranking between $m$ perturbed versions of $\rho_k$.  The perturbed versions of the prompt are obtained as $\rho_k' = \rho_k + \epsilon \mathcal{N}(0, I_d)$, where $\epsilon$ is the noise scale, $I_d$ is the $d \times d$ identity matrix, and $d = |\rho|$ is the length of the prompt.
The prompt is then updated according to $\rho_{k+1} \leftarrow \rho_k + \eta\hat{\nabla}_{\rho} G$, where $\eta$ is the learning rate that we anneal from $1$ to $0.1$ over the $K$ rounds. 
Crucially, at each round $k$, all of the $m$ prompt-perturbations must be evaluated with an online rollout of the PDT, meaning the sample complexity of this method is $m$ times larger than that of our method.

\textbf{Gaussian perturbation hill climbing}: A simple stochastic optimization approach inspired by hill climbing. Given an initial prompt $\rho \sim \mathcal{P}_i$, we iteratively perturb the sampled prompt by applying Gaussian noise. 
At each round $k$, the perturbed prompt is obtained as $\rho_k = \rho + \epsilon \mathcal{N}(0, I_d)$.
We anneal $\epsilon$ from $1$ to $0.1$ over the $K$ rounds. The perturbed prompt $\rho_k$ is evaluated by rolling out the PDT. If the resulting return $G_k$ exceeds the best return so far, we update the prompt $\rho \leftarrow \rho_k$, thereby performing hill climbing with respect to online task return directly in the prompt space.

\subsection{Results \& Analysis}\label{sec:results}
\textbf{Does bandit-based prompt-tuning improve a frozen PDT backbone}?
We perform prompt tuning on the pre-trained PDT $\theta^*$ using 250 online rollouts on training tasks with radius $r=2.9$. We run this experiment with $J \in \{ 1, 2, 4\}$, i.e., with increasingly many segments and, conversely, tokens for task identification in the prompt. We run our bandit-based prompt-tuning method with UCB~\cite{li2010contextual} and $\epsilon$-greedy exploration strategies.

\begin{figure*}[t!]
    \centering
    \includegraphics[width=0.7\textwidth]{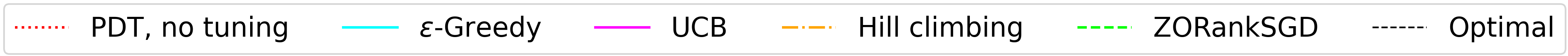}
    \begin{subfigure}[t]{0.3\textwidth}
        \centering
        \includegraphics[width=\linewidth]{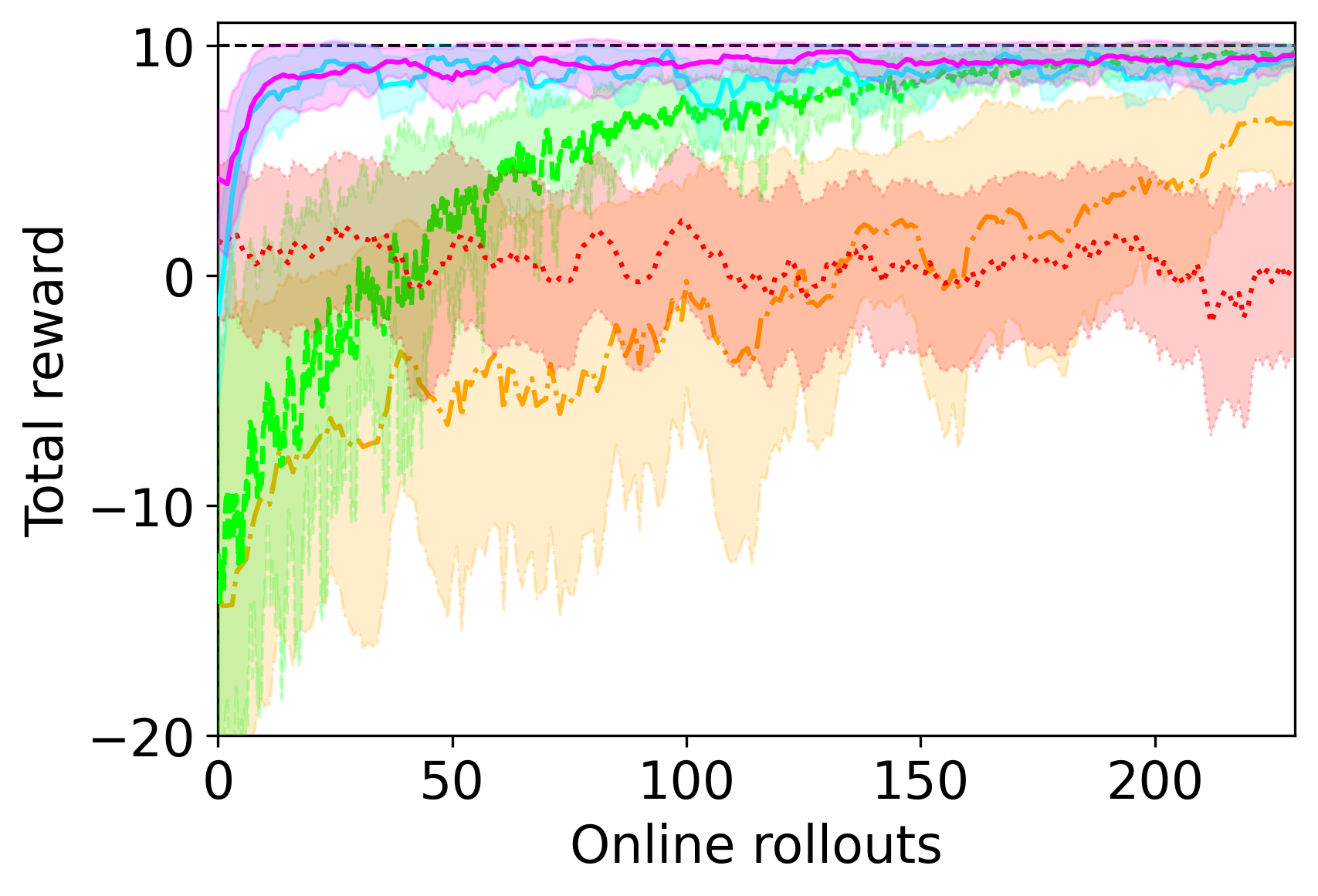}
        \caption{$J=1$}\label{subfig:online_j1}
    \end{subfigure}
    ~
    \begin{subfigure}[t]{0.3\textwidth}
        \centering
        \includegraphics[width=\linewidth]{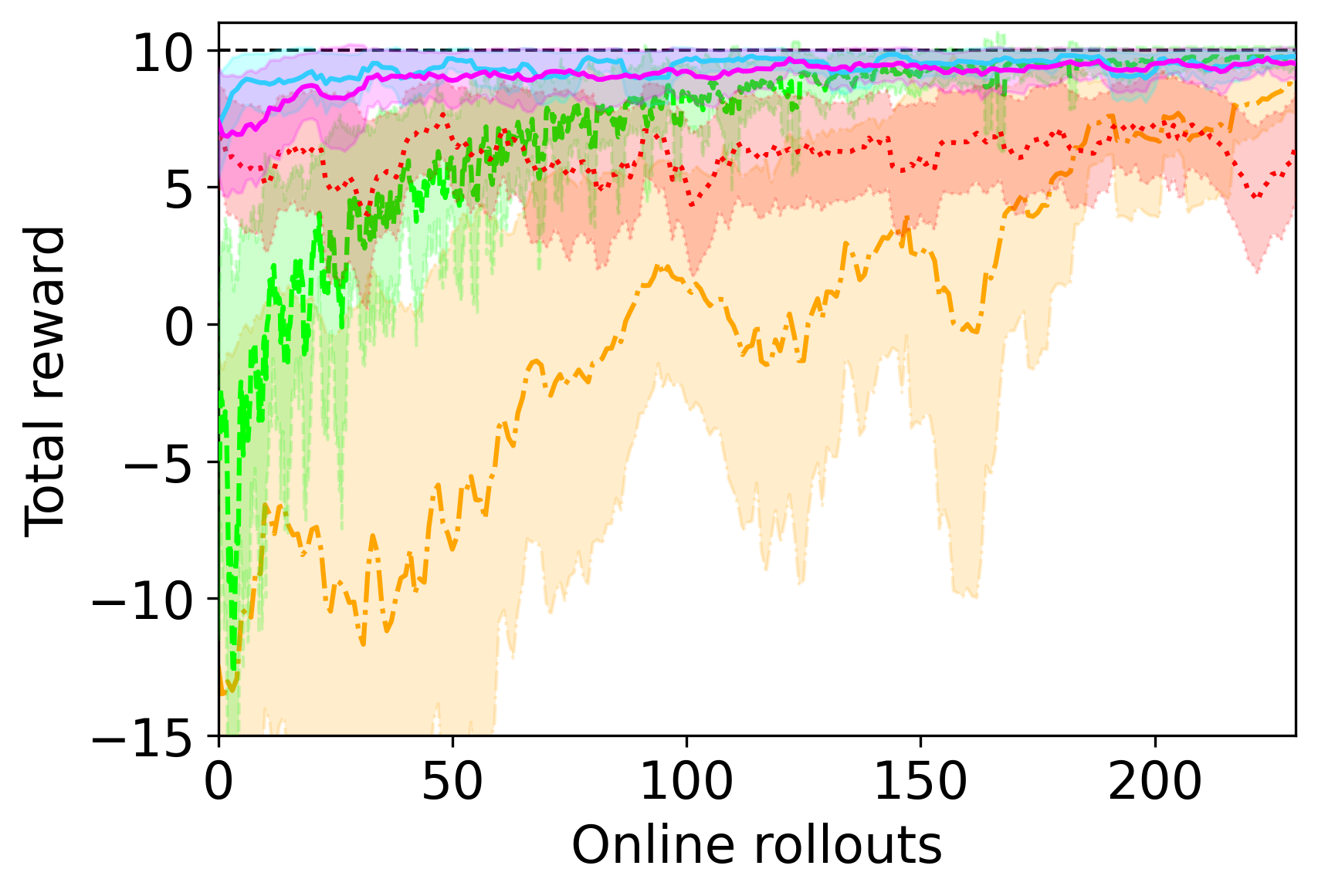}
        \caption{$J=2$}\label{subfig:online_j2}
    \end{subfigure}
    ~
    \begin{subfigure}[t]{0.3\textwidth}
        \centering
        \includegraphics[width=\linewidth]{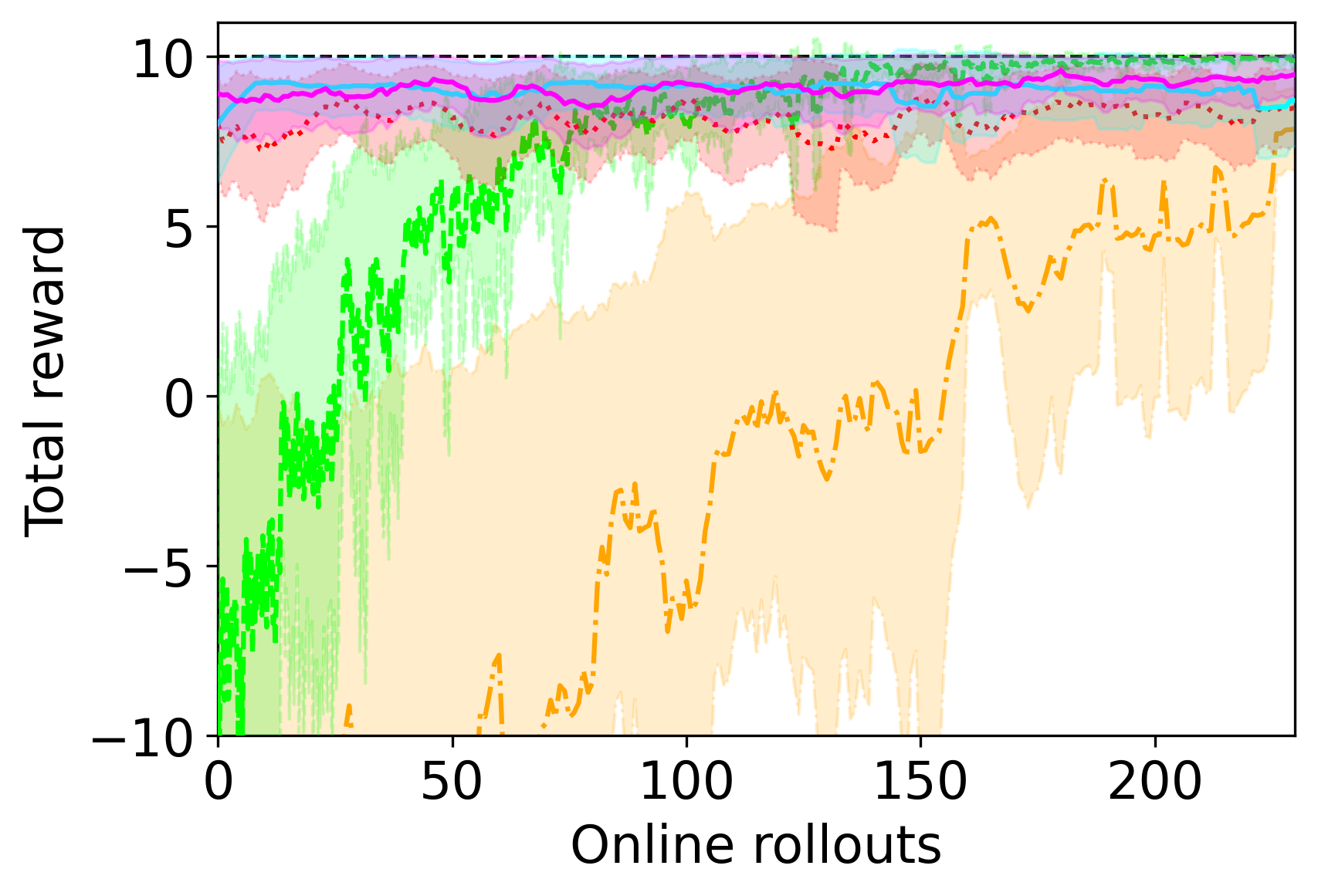}
        \caption{$J=4$}\label{subfig:online_j4}
    \end{subfigure}
    \caption{Inference time performance gains due to prompt-tuning over 250 episodes. 
    ZO-RankSGD performs a total of $m=5 \times 250$ rollouts 
    to estimate the prompt gradient 250 times, which we squash into the 0 - 250 range in the plot.
    Results are averaged over all training tasks and three seeds. To increase readability, the shaded area corresponds only to 0.25 standard deviations around the mean.}\label{fig:online-propmt-tuning}
\end{figure*}

Results are shown in Fig.~\ref{fig:online-propmt-tuning}, performance gains due to prompt-tuning are most prominently visible in Fig.~\ref{subfig:online_j1}, where the prompt consists of a single segment of length $H=3$, for a total of $J\times H \times (|\mathcal{S}| + |\mathcal{A}| + 1) = 1 \times 3 \times (2 + 3 + 1) = 18$ prompt tokens.
Despite being trained to convergence on all training tasks, PDT without prompt tuning fails to achieve the optimal return. This shortfall is due to the uninformed, random prompt sampling strategy used by standard PDT which frequently selects uninformative prompts, limiting its performance.
Our bandit-based prompt-tuning approach, however, quickly boosts the performance of the underlying PDT backbone to optimal levels of return by identifying high-return prompts, with no considerable difference between $\epsilon$-greedy or UCB exploration.
The other prompt-tuning baseline methods, Gaussian perturbation with hill climbing and ZO-RankSGD-based prompt-tuning, also demonstrate clear improvements over the course of the 250 online rollouts, though they are less efficient than the bandit approach. Notably, ZO-RankSGD requires $m=5$ additional online rollouts for each prompt-gradient estimation, resulting in a total of $5 \times 250 $ online rollouts. In contrast, our bandit-based approach rapidly converges,  consistently selecting optimal prompts within the first few rollouts. 

Additionally, we observe the following trends as prompt size increases.
First, although PDT performance without prompt-tuning remains roughly constant over 250 rollouts, it scales approximately proportionally with prompt size, reducing the performance gain from prompt-tuning.
In Fig.~\ref{subfig:online_j4}, PDT achieves near-optimal return even without prompt-tuning, which implies that, in our proof-of-concept environment, exhaustive random sampling suffices for finding tokens that uniquely identify the downstream task.
Interestingly, both Gaussian perturbation and the ZO-RankSGD baseline scale poorly with the prompt size.
We hypothesize that this stems from their strategy of perturbing the \textit{entire} prompt at each round, which can unnecessarily disrupt informative segments by injecting excessive noise, even when the original prompt is nearly optimal.
In contrast, our bandit-based method avoids this issue by exploring prompt segments independently with each arm. This enables it to preserve high-performing segments while selectively exploring others, \textit{without} adding unnecessary noise to effective parts of the prompt.

\begin{figure*}[t]
    \centering
    \begin{subfigure}[t]{0.45\textwidth}
        \centering
        \includegraphics[width=\linewidth]{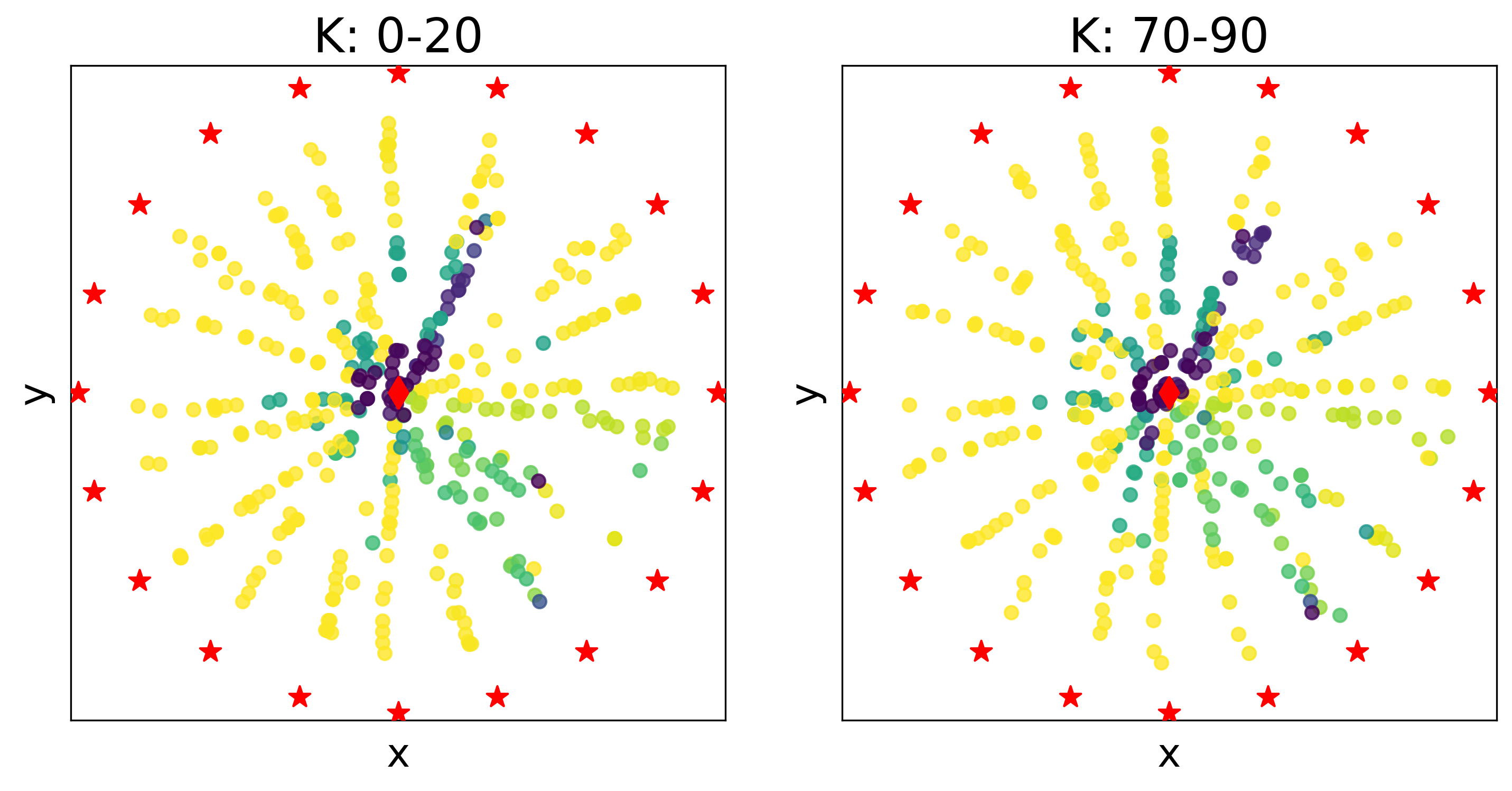}
        \caption{Standard PDT, no tuning}\label{subfig:spatio-random}
    \end{subfigure}
    ~
    \begin{subfigure}[t]{0.45\textwidth}
        \centering
        \includegraphics[width=\linewidth]{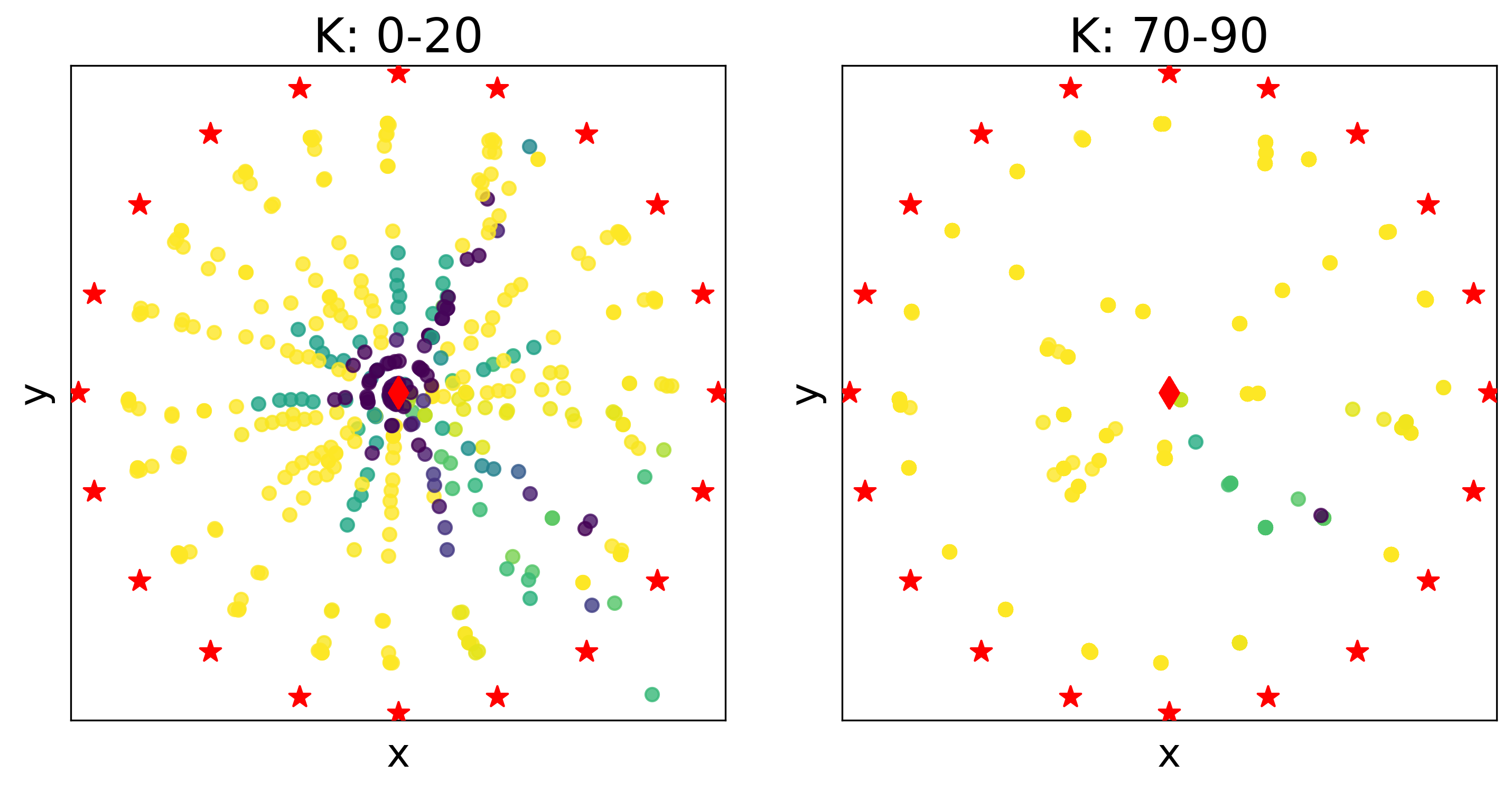}
        \caption{Our CMAB approach, $\epsilon$-greedy}\label{subfig:spatio-bandit}
    \end{subfigure}
    ~
    \begin{subfigure}[t]{0.45\textwidth}
        \centering
        \includegraphics[width=\linewidth]{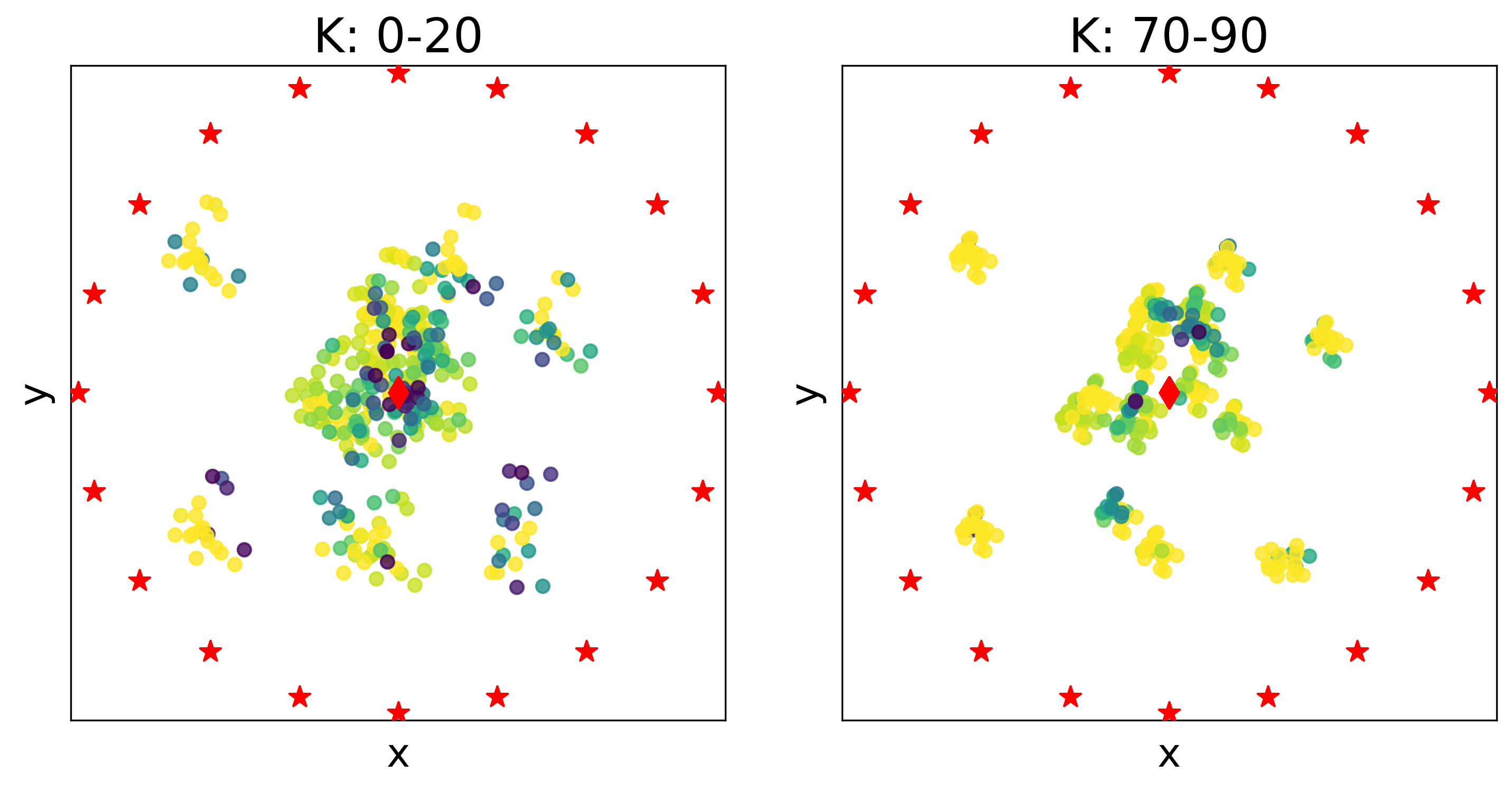}
        \caption{Gaussian perturbation and hill climbing}\label{subfig:spatio-gaussian}
    \end{subfigure}
    ~
    \begin{subfigure}[t]{0.45\textwidth}
        \centering
        \includegraphics[width=\linewidth]{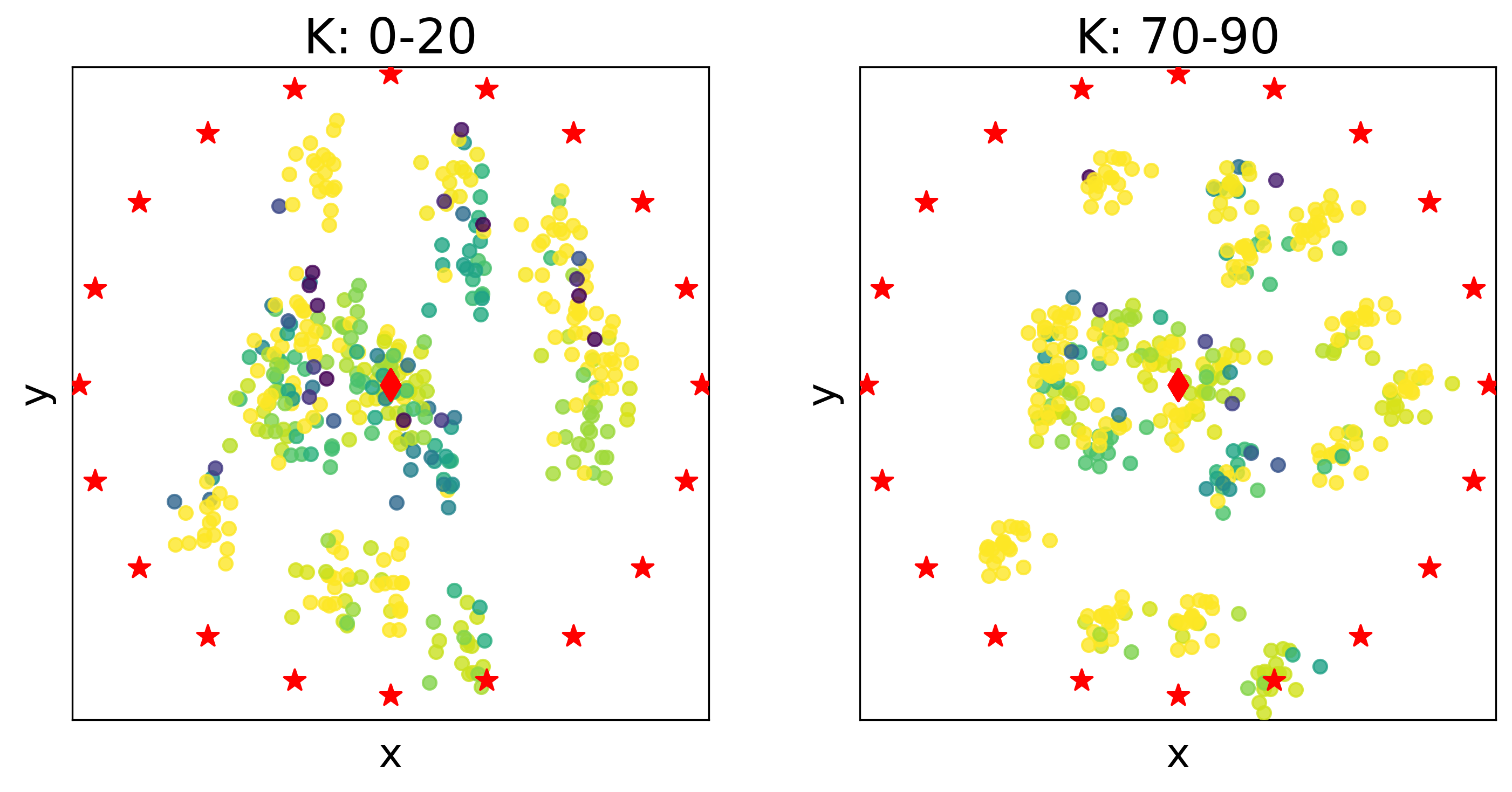}
        \caption{ZO-RankSGD \cite{xu2022prompting}}\label{subfig:spatio-zoranksgd}
    \end{subfigure}
    ~
    \caption{Spatio-temporal comparison between prompt selection approaches. Prompts are plotted by the mean spatial coordinate of the states in the prompt and colored according to the achieved return \includegraphics[height=\baselineskip]{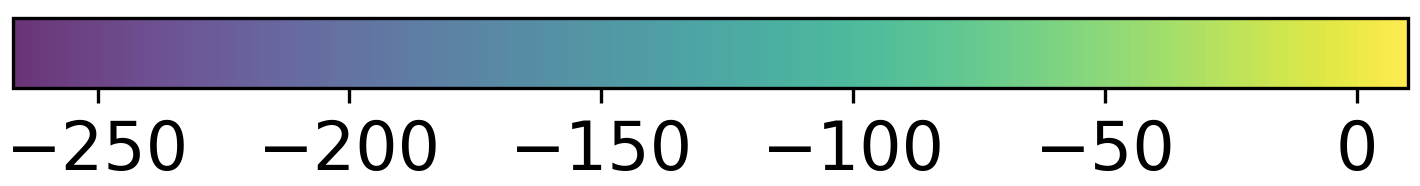} when using that prompt. The MDP's starting state is indicated with the red diamond, and the goal states for different tasks are indicated by the red stars. $K$ denotes the bandit rounds for each image.}\label{fig:spatio-bandit-gaussian}
\end{figure*}

\textbf{How does bandit-based prompt-tuning explore the prompt space?}
We visualize selected prompts in the beginning ($K$: 0 - 20) and towards later stages ($K$: 70 - 90) of exploration in Fig~\ref{fig:spatio-bandit-gaussian}. 
PDT's uniform prompt selection strategy, with no difference between the early or later stages, can be seen in Fig.~\ref{subfig:spatio-random}.
As shown in Fig.~\ref{subfig:spatio-bandit}, our bandit-based approach initially explores the entire prompt space, experiencing low- and high-performance prompts. However, in later rounds, the bandit prioritizes prompts that are closer to the goal states while avoiding the low-performance prompts near the center,  as these provide less informative signals for the task.

The Gaussian perturbation method in Fig.~\ref{subfig:spatio-gaussian} primarily explores locally. This is due to the hill climbing optimization, which finds the best-performing prompt in vicinity of the initially sampled prompt while falling short of exploring the whole prompt space. 
The ZO-RankSGD-based prompt-tuning in Fig.~\ref{subfig:spatio-zoranksgd} similarly explore only locally, revealing a strong dependence on the initialization. 
Unlike incremental approaches, our bandit method is less reliant on the initially sampled prompt. Instead, it exploits segment similarities to identify the best segments in $\mathcal{P}_i$.  
This result highlights a key limitation of perturbation-based methods in prompt-space exploration and illustrates how our bandit approach effectively selects prompts that drive performance improvements.

\section{Conclusion \& Future Work}
We introduce a bandit-based prompt-tuning approach that enables few-shot generalization at inference time for multi-task decision transformers. By efficiently exploring the prompt space, our approach identifies high-performing prompts for downstream tasks, avoiding exhaustive random sampling or full-model finetuning.

Our preliminary results show that the proposed contextual bandit architecture efficiently boosts a pre-trained PDT to optimal levels of performance, after only a small number of online rollouts, while uniform sampling and prior prompt-tuning approaches underperform.

We believe these findings warrant future research and are exploring the scalability of the proposed method to more complex and challenging environments and aim to present additional results at the workshop.

\bibliography{main}

\begin{thebibliography}{10}
\providecommand{\url}[1]{\texttt{#1}}
\providecommand{\urlprefix}{URL }
\providecommand{\doi}[1]{https://doi.org/#1}

\bibitem{auer2002finite}
Auer, P., Cesa-Bianchi, N., Fischer, P.: Finite-time analysis of the multiarmed bandit problem. Mach. Learn.  \textbf{47}(2–3),  235–256 (May 2002). \doi{10.1023/A:1013689704352}, \url{https://doi.org/10.1023/A:1013689704352}

\bibitem{brown2020language}
Brown, T., Mann, B., Ryder, N., Subbiah, M., Kaplan, J.D., Dhariwal, P., Neelakantan, A., Shyam, P., Sastry, G., Askell, A., et~al.: Language models are few-shot learners. Advances in neural information processing systems  \textbf{33},  1877--1901 (2020)

\bibitem{pmlr-v235-chen24e}
Chen, L., Chen, J., Goldstein, T., Huang, H., Zhou, T.: {I}nstruct{Z}ero: Efficient instruction optimization for black-box large language models. In: Proceedings of the 41st International Conference on Machine Learning. vol.~235, pp. 6503--6518 (2024)

\bibitem{chen2021decision}
Chen, L., Lu, K., Rajeswaran, A., Lee, K., Grover, A., Laskin, M., Abbeel, P., Srinivas, A., Mordatch, I.: Decision transformer: Reinforcement learning via sequence modeling. Advances in neural information processing systems  \textbf{34},  15084--15097 (2021)

\bibitem{dosovitskiy2020image}
Dosovitskiy, A.: An image is worth 16x16 words: Transformers for image recognition at scale. arXiv preprint arXiv:2010.11929  (2020)

\bibitem{hu2023prompt}
Hu, S., Shen, L., Zhang, Y., Tao, D.: Prompt-tuning decision transformer with preference ranking. arXiv preprint arXiv:2305.09648  (2023)

\bibitem{hu2024prompt}
Hu, S., Zhao, W., Lin, W., Shen, L., Zhang, Y., Tao, D.: Prompt tuning with diffusion for few-shot pre-trained policy generalization. arXiv preprint arXiv:2411.01168  (2024)

\bibitem{lester2021power}
Lester, B., Al-Rfou, R., Constant, N.: The power of scale for parameter-efficient prompt tuning. arXiv preprint arXiv:2104.08691  (2021)

\bibitem{li2010contextual}
Li, L., Chu, W., Langford, J., Schapire, R.E.: A contextual-bandit approach to personalized news article recommendation. In: Proceedings of the 19th international conference on World wide web. pp. 661--670 (2010)

\bibitem{li2023survey}
Li, W., Luo, H., Lin, Z., Zhang, C., Lu, Z., Ye, D.: A survey on transformers in reinforcement learning. arXiv preprint arXiv:2301.03044  (2023)

\bibitem{lin2023use}
Lin, X., Wu, Z., Dai, Z., Hu, W., Shu, Y., Ng, S.K., Jaillet, P., Low, B.K.H.: Use your instinct: Instruction optimization using neural bandits coupled with transformers. In: NeurIPS 2023 Workshop on Instruction Tuning and Instruction Following (2023)

\bibitem{radford2018improving}
Radford, A.: Improving language understanding by generative pre-training  (2018)

\bibitem{radford2021learning}
Radford, A., Kim, J.W., Hallacy, C., Ramesh, A., Goh, G., Agarwal, S., Sastry, G., Askell, A., Mishkin, P., Clark, J., et~al.: Learning transferable visual models from natural language supervision. In: International conference on machine learning. pp. 8748--8763. PMLR (2021)

\bibitem{reed2022generalist}
Reed, S., Zolna, K., Parisotto, E., Colmenarejo, S.G., Novikov, A., Barth-maron, G., Gim{\'e}nez, M., Sulsky, Y., Kay, J., Springenberg, J.T., et~al.: A generalist agent. Transactions on Machine Learning Research

\bibitem{schulman2017proximal}
Schulman, J., Wolski, F., Dhariwal, P., Radford, A., Klimov, O.: Proximal policy optimization algorithms. arXiv preprint arXiv:1707.06347  (2017)

\bibitem{shi2024best}
Shi, C., Yang, K., Yang, J., Shen, C.: Best arm identification for prompt learning under a limited budget. In: ICLR 2024 Workshop on Understanding of Foundation Model (2024)

\bibitem{tang2023zeroth}
Tang, Z., Rybin, D., Chang, T.H.: Zeroth-order optimization meets human feedback: Provable learning via ranking oracles. arXiv preprint arXiv:2303.03751  (2023)

\bibitem{thompson1933likelihood}
Thompson, W.R.: On the likelihood that one unknown probability exceeds another in view of the evidence of two samples. Biometrika  \textbf{25}(3/4),  285--294 (1933)

\bibitem{xu2022prompting}
Xu, M., Shen, Y., Zhang, S., Lu, Y., Zhao, D., Tenenbaum, J., Gan, C.: Prompting decision transformer for few-shot policy generalization. In: international conference on machine learning. pp. 24631--24645. PMLR (2022)

\bibitem{yuan2024pre}
Yuan, H., Fu, Y., Xie, F., Lu, Z.: Pre-trained multi-goal transformers with prompt optimization for efficient online adaptation. Advances in Neural Information Processing Systems  \textbf{37},  55086--55114 (2024)

\end{thebibliography}

\end{document}